\documentclass[11pt]{article}

\usepackage[final]{acl}

\usepackage{times}
\usepackage{latexsym}
\usepackage{amsmath}
\usepackage{amssymb}
\usepackage[T1]{fontenc}

\usepackage[utf8]{inputenc}

\usepackage{microtype}

\usepackage{inconsolata}

\usepackage{graphicx}

\title{Multi-author style detection with stylistic embeddings}

\author{Man-Ling Sung, Chak Fai Li, Damianos Karakos, Fred Bernardin, Brian Ulicny 
 \\
 RTX BBN Technologies \\ Cambridge MA, USA
 }

\begin{document}
\maketitle
\begin{abstract}

We investigate the use of Large Language Models (LLMs) in multi-author style change detection. Using off-the-shelf LLMs or supervised fine-tuning solely within the target domain fails to yield effective features; for this reason, we propose cross-domain transfer learning with LLM-based embeddings fine-tuned for authorship verification. We fine-tune Mistral-7B embedding model and compare it with DeBERTa, used in the state-of-the-art (SOTA) systems in PAN-25 
multi-author writing style analysis challenge.
Results show that DeBERTa and Mistral excel in different user-generated text scenarios. We exploit this diversity by combining both embeddings, leading to more effective style change detection.
\end{abstract}

\section{Introduction}

In the domain of user-generated text, authorship detection has long been an important task for verifying the consistency of a user across the text they generate online. Spotting changes in authorship is important for detecting security incidents, such as account hijacking/takeover or plagiarism. In such cases, it is important to detect if multiple users contributed to the writing: the hijacker and the plagiarist may only change {\em parts} of a document, not the entire document, to evade detection. This problem becomes much harder for online text due to typos, slang, and acronyms. Despite recent advancements in Large Language Models (LLMs), and despite their popularity and success in tackling natural language tasks, they still struggle to directly serve as feature extractors that can effectively capture authorship information. In the PAN-25 multi-author writing style analysis task, DeBERTa (an encoder-based transformer) still performs better as the stylistic embeddings than LLMs -- a decoder-based model. 
However, there are successful attempts at adopting LLMs for cross-domain authorship attribution by extracting stylistic information. One adaptation framework, which we use in our work, is {\em Learning Universal Authorship Representations} (LUAR) \cite{rivera2021learning}.

Our work explores the use of LLM-based stylistic embeddings for multi-author detection. We want to understand whether \textbf{cross-domain authorship representations, learned with a decoder-based embedding model, capture stylistic signals that are complementary to the contextual signals captured by DeBERTa}. We propose a resource-efficient LLM adapted system for effective sentence-level style change detection. We also analyze different types of user-generated text and propose a more versatile system for effective style change detection leveraging both encoder-based and decoder-based models.

\section{Problem Definition and Related Work}

\subsection{Multi-author style change detection}

Given a multi-author document, the goal of style change detection is to identify the text positions where the author switches within the document. This task is a form of intrinsic authorship analysis that does not rely on any additional text sources besides the single given document during detection.

We focus on author change detection at the sentence level. For example, in a two-author document $D = \{d_1, d_2, ... , d_N\}$ with $N$ sentences and with a single transition point, the goal is to identify the transition sentence $\hat{i}$ with the property that $\{d_1, .. , d_{\hat{i}-1}\}$ is written by author A and $\{d_{\hat{i}}, .. , d_N\}$ is written by author B. In general multi-author documents, the goal is to detect all transition sentences where their adjacent sentences are written by different authors.

\subsection{Stylistic feature representations}
The most common approach to style change detection is the ensemble technique, which embeds a text sentence into representation features for a decision-making model. Decision-making models can range from simple cosine similarity comparison to machine learning or neural network (NN) classifiers. 
For representation features, classic methods are based on stylometry  \cite{evert2017understanding}, 
such as Bag-of-Words (BoW) \cite{bevendorff2025overview}, or word graph with carefully selected linguistic features \cite{boriceanu2025style}. Current state-of-the-art (SOTA) uses encoder-only transformer models, which are designed to understand contextual meaning of the text, specifically with DeBERTa \cite{he2021DeBERTav3}. 
In PAN-25 multi-author writing analysis task, the two best performing systems both utilize DeBERTa. The first team \cite{lin2025team} extracts sentence features with DeBERTa, then trains a NN classifier. Another team \cite{lin2025scl} uses supervised contrastive learning with joint optimization of classification and contrastive loss, using DeBERTa as the underlying language model. This reflects DeBERTa's strong capability to capture distinguishable stylistic features that a simple classifier can easily learn. 

Despite the recent emergence of LLMs,  attempts to use pretrained LLMs have not yielded strong results. Llama-3-8B works well at paragraph level \cite{lv2024team}, but struggles at sentence level \cite{chen2025llama}. In sentence level detection, whether it is trying to extract sentence representation with pre-trained LLM, e.g. Claude \cite{schmidt2025team}, or fine-tuned Llama-3, both struggle to see competitive improvement beyond DeBERTa.  One limitation on fine-tuning LLM is the vast amount of data required to effectively adapt the model to the target domain. As far as we know, there is no prior work in using existing LLM embeddings for such a task.





\subsection{LLM adaptation for stylistic embeddings}

With limited in-domain data for model adaptation, the focus has been on cross-domain transfer. In authorship verification, a universal framework -- Learning Universal Authorship Representations (LUAR) -- is proposed for effective domain transfer \cite{rivera2021learning}. It adopts a contrastive-learning framework to produce style-focused embeddings on top of a language model text embeddings, with max-over-time pooling, which is robust across topics and domains. The stylistic features trained on human authors, can also be transferred to detect machine-generated text from human \cite{rivera2024few, kandula2025bbn}.

Efforts have also been made to enable an LLM to perform specific tasks where it normally fails. This approach typically relies on carefully crafted, task-specific data to fine-tune the model. For instance, cross-domain authorship attribution with LLMs can be framed as a retrieval task \cite{agarwal2025cross}. Along these lines, 
Linq-embed-mistral \cite{choi2024linq} is a fine-tuned version of a Mistral model specifically for text retrieval. 

\section{Methodology}

\begin{figure*}  \centering
  \includegraphics[width=0.6\paperwidth]{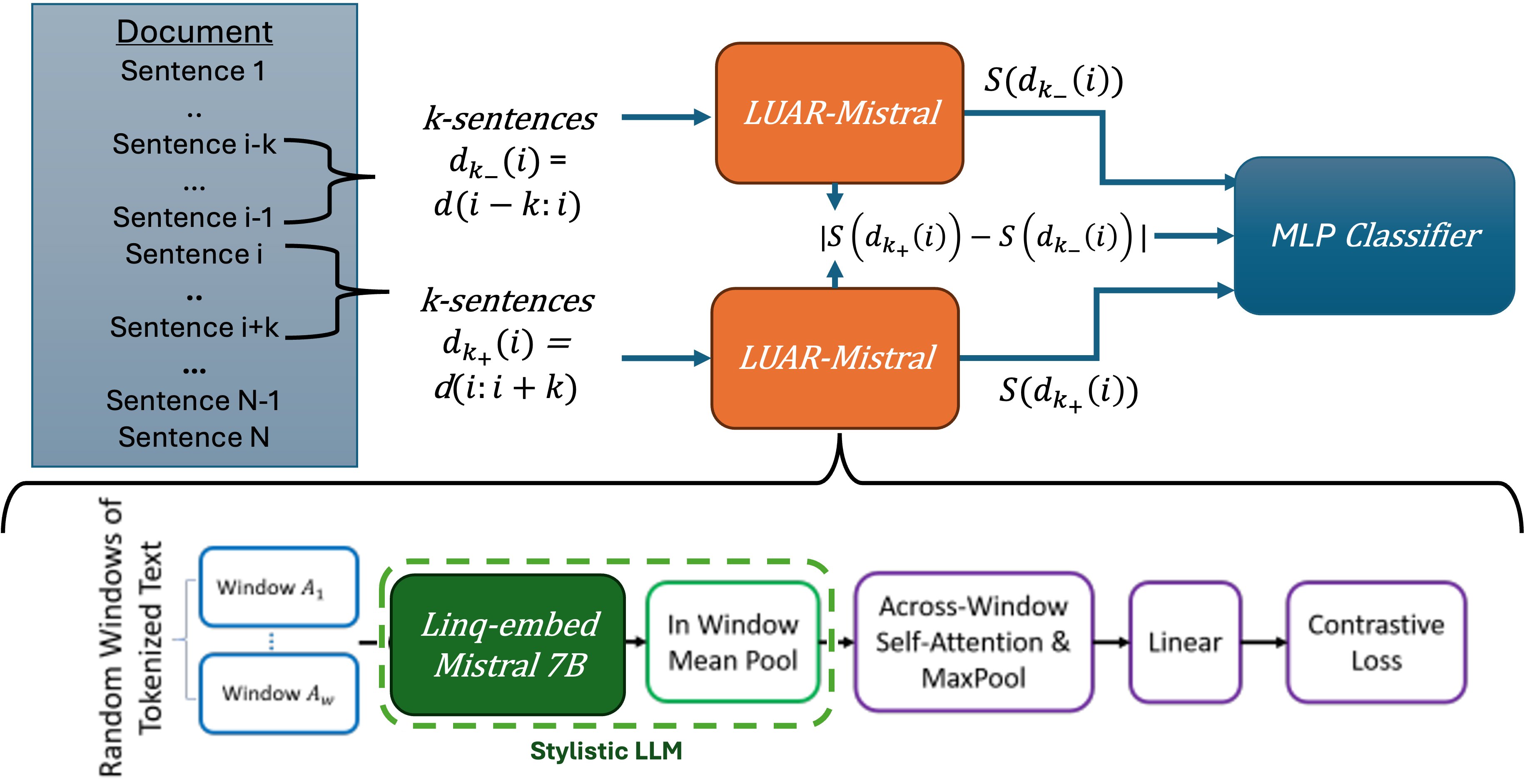}
  \caption{The proposed system. LUAR-Mistral generates embeddings from k sentence-pairs, where k is the number of sentences. This figure illustrated the single sentence case where k=1. The MLP classifier takes in both sentence embeddings and their absolute difference to decide if authorship change happens in between. The sentence embeddings come from a modified LUAR framework which replaces BERT with Linq-embeded Mistral. The framework is fine-tuned to optimize the contrastive loss of same and different authors.}
   \label{fig:system_design}
\end{figure*}

Our system performs two steps, a LLM is used for feature extraction, followed by sentence-level classification with a neural network classifier. The system diagram is presented in Figure \ref{fig:system_design}. 

\subsection{Feature Extraction Model}

We adopted a training methodology in \citealp{kandula2025bbn}, a LUAR framework that fine-tunes the Linq-embed-mistral model with contrastive learning. Mistral-7B is fine-tuned with LoRA and low-bit quantization. 
To extend its capability, the model is further fine-tuned on open-source datasets to comprehensively cover multiple aspects of user-generated text. Datasets span diverse domains, including social media discussions, reviews and comments, email and narratives (see Appendix \ref{sec:app_decision} for details). 
We also explored replacing BERT to other white-box models in LUAR, such as Qwen2 \cite{qwen2.5} and multilingual-e5-large (mE5) \cite{wang2024multilingual}. Linq-embed-mistral (LE-Mistral-7B) performs the best on all PAN-25 tasks, as can be seen in Table \ref{tab:lura_models}. 

\begin{table}[h]
\begin{tabular}{llll} \hline
PAN-25 (Valid) & Easy   & Medium & Hard   \\ \hline
Qwen2-1.5B    & 0.872 & 0.798  & 0.703 \\
LE-Mistral-7B & \textbf{0.904} & \textbf{0.834}  & \textbf{0.734} \\
mE5     & 0.862 & 0.772  & 0.685 \\\hline
\end{tabular}
\caption{Results on PAN-25 validation set adopted with LUAR framework. Sentence-level features is used. For an explanation of the easy/medium/hard conditions, please read Section \ref{sec:dataset}.}
\label{tab:lura_models}
\end{table}

\subsection{Neural Network Classifier}
We train a binary classifier on the sentence embeddings along with their absolute difference. The architecture consists of fully connected multi-layer perceptron (MLP) layers, with ReLU activation and a dropout layer of rate 0.1. It is trained with batch size 64, learning rate $1 \times 10^{-5}$, AdamW optimizer and BCEWithLogitsLoss. 

The classifier, in its baseline form, uses pairs of adjacent sentence features as input, as done in the systems submitted in PAN-25. However, we believe that incorporating additional context facilitates detection, especially in short sentences with limited stylistic information (commonly found on social media). We therefore explore a second feature variation, which includes additional features spanning $2k$ sentences, separated in the middle by the hypothesized transition sentence. In the first case (just pairs of adjacent sentences) we use a 2-layer MLP, presented in equation (1); in the second case ($k$ + $k$ sentences) we use a 3-layer MLP, presented in equation (2). The second approach integrates structural information, enabling the model to look beyond single sentences. The decision of the classier $\hat{y}$, is computed as follows for each MLP classifier.

The 2-layer MLP that takes in a pair of sentence embeddings and their absolute difference:
\begin{align}
\hat{y} = \sigma(\mathbf{W_2} \cdot Dropout( ReLU(\mathbf{W_1} \cdot \mathbf{S}_{LM}))
\end{align}
where $\mathbf{W_2} \in \mathbb{R}^{1\times0.5f}$,  $\mathbf{W_1} \in \mathbb{R}^{0.5f\times3f}$ are learnable parameters. $\mathbf{S}_{LM}$ denotes the input embedding from the fine-tuned language model and $f$ is the embedding size. In the case of Linq-embed Mistral 7B, $f=1536$, and in the case of Qwen2 and multilingual-e5-large, $f=1024$.



The 3-layer MLP, which on top of the features fed into the 2-layer 
MLP, takes in an additional set of multi-sentence embedding pairs and their difference:
\begin{multline}
\hat{y} = \sigma(\mathbf{W^{\prime}_3} \cdot Dropout(ReLU (\mathbf{W^{\prime}_2} \cdot \\Dropout(ReLU(\mathbf{W^{\prime}_1} \cdot \mathbf{S}_{LM})))) 
\end{multline}

where $\mathbf{W^{\prime}_3} \in \mathbb{R}^{1\times0.5f}$,  $\mathbf{W^{\prime}_2} \in \mathbb{R}^{0.5f\times3f}$, $\mathbf{W^{\prime}_1} \in \mathbb{R}^{3f\times6f}$  are learnable parameters. $\mathbf{S}_{LM}$ and $f$ are described above.


\section{Experiments and Analysis}

\subsection{Dataset}
\label{sec:dataset}




We use the PAN-25/26 multi-author writing style analysis datasets. This problem focuses on detecting transition sentences where the change of authorship happens. 
PAN-25 \cite{bevendorff2025overview} combines Reddit comments into documents. The dataset includes discussions about world news, politics, history, and legal advice. PAN-26 \cite{bevendorff2026overview} contains
user-written stories that reuse fictional characters and settings from existing works. Both datasets are split into 3 tasks of various difficulty levels -- easy, medium and hard.
For the easy task, each document contains multiple topics written by different authors, allowing detection to be framed around identifying topic changes. For the medium task, documents are topically coherent, while stylistic similarity between sentences is moderate. For the hard task, the documents contain a single topic from various authors who are stylistically similar. Our primary interest lies in the medium and hard tasks, with a focus on detecting stylistic differences within the same topic. Notably, PAN-26 is much larger than PAN-25, with 10 times more sentences and 2-3 times more documents. Detailed analysis is in Table \ref{tab:dataset_info}. 

\subsection{Experiment Design}
\label{sec:exp_design}

The classifiers are trained on PAN training sets for 20 epochs and scored against the PAN validation sets with decision threshold tuning to maximize the F1-score. The model checkpoint that gives the highest F1 on validation set is used for evaluation. 
We evaluated three versions of our system: 1) the baseline Linq-Embed-Mistral-7B, 2) \textbf{LUAR-Mistral}: Linq-Embed-Mistral-7B fine-tuned with public data (shown in Appendix \ref{sec:app_decision}) using LUAR, 3) \textbf{LUAR-Mistral-PAN}: LUAR-Mistral further fine-tuned with provided PAN training set. We use DeBERTa-v3a to cluster the sentences from the target training set, and selected 40k sentences uniformly from the clusters to create sentences pairs for LUAR.
We extract the features at sentence-level and multi-sentence-level, subsequently fitting them to their respective classifiers. For multi-sentence-level setup we sweep the number of sentences $k = [ 2 ... 7 ]$. 

We reproduce the challenge baseline (BoW with cosine similarity), as well as DeBERTa-Classifier and SCL-DeBERTa. We did not reproduce the LLM-based systems from PAN-25 due to their overlapping nature with the Mistral model evaluated in this work. Both the DeBERTa and Mistral systems were executed on a single NVIDIA A6000 GPU.

\subsection{Results and Analysis on PAN-25/26}

\begin{table*}[h]
\centering

\begin{tabular}{lllllll}\hline
\textbf{Task}                              & \multicolumn{3}{c}{\textbf{PAN 2025 (test)}}      & \multicolumn{3}{c}{\textbf{PAN 2026 (valid)}}     \\ \hline 
\textbf{Model}                                                    & \textbf{Easy}  & \textbf{Medium} & \textbf{Hard}  & \textbf{Easy}  & \textbf{Medium} & \textbf{Hard}  \\ \hline
Baseline  \cite{bevendorff2025overview}                                           & 0.439          & 0.44            & 0.453          & 0.410          & 0.489           & 0.442          \\
Llama-3 (4-bit) ft \cite{chen2025llama} & 0.461          & 0.583           & 0.484          & /              & /               & /              \\
Claude-3.7-sonnet \cite{schmidt2025team}                                                & 0.856          & 0.818           & 0.661          & /              & /               & /              \\
SCL-DeBERTa \cite{lin2025scl}                                          & 0.950          & 0.812           & 0.827          & 0.980          & 0.550           & 0.489          \\
DeBERTa-Classifier \cite{lin2025team}   & \textbf{\underline{0.958}} & 0.822           & \underline{0.829} & \textbf{\underline{1.000}} & 0.633           & 0.735          \\ \hline
Linq-Embed-Mistral-7B  \cite{choi2024linq}       & 0.928          & 0.790           & 0.731 &   0.997    &   0.613   & 0.676     \\
LUAR-Mistral                & 0.938          & 0.830           & 0.762          & 0.992          & 0.682           & 0.740          \\
\quad + k sentences               & 0.946          & \underline{0.845}  & 0.772          & 0.992          & \underline{0.707}  & \underline{0.743} \\
LUAR-Mistral-PAN            & 0.936          & 0.830           & 0.763          & 0.990          & 0.650           & 0.736          \\
\quad + k sentences               & 0.947          & 0.844           & 0.766          & 0.992          & 0.702           & 0.741          \\\hline
\multicolumn{7}{c}{\textbf{System Combination (Deberta+LUAR-Mistral)}}                                                              \\\hline
Deberta + LUAR-Mistral      & 0.952          & 0.846           & 0.828          & \textbf{1.000} & 0.689           & 0.764          \\
\quad + k sentences               & \textbf{0.958} & \textbf{0.858}  & \textbf{0.844} & \textbf{1.000} & \textbf{0.718}  & \textbf{0.769} \\ \hline
\end{tabular}
\caption{Results on PAN-25/26. The top table contains results with previous systems, second table is with our own systems, bottom table is with combined systems. The results for LLama-3 and Claude-3.7-sonnet are copied from \cite{zangerle2025overview}, while the rest are our own reproduction. The best overall result in each column is bolded, while the best single system result is underlined.}
\label{tab:pan2526_results}
\end{table*}

We follow the PAN challenge that measures the performance in Macro-F1 score, the results are presented in Table \ref{tab:pan2526_results}. DeBERTa achieves the highest performance on PAN-25 easy/hard tasks and PAN-26 easy task, while the Mistral systems excel in PAN-25 medium and PAN-26 medium/hard tasks. Extending neighboring sentences yields better performance, especially for PAN-25 hard and PAN-26 medium tasks. The gain from further fine-tuning LUAR-Mistral with carefully crafted domain data is minimal compared to increasing the context window for the sentence embeddings.

Surprisingly, even though the PAN-26 training set is much larger than PAN-25, detection accuracy drops significantly on medium/hard tasks. The systems struggle to detect authorship change even with more training data. We suspect the system can learn to detect transition words instead of the semantic or stylistic difference within sentences in PAN-25 to yield such high accuracy (see examples in Appendix \ref{sec:appendix_dataset}).
It is known that transformer based models, such as DeBERTa perform better than LLMs in disentangling detection \cite{azuma2025svm}.

\section{Comparing DeBERTa and Mistral}
\subsection{Experiment on HIATUS Research Set}

To validate our hypothesis, we run the same multi-author detection task on the HIATUS Research Datasets (HRS) \cite{iarpa2024hiatus}. It consists of five genres - 
board-game reviews, instructions, journalism, posts from literature and STEM forums. 
We construct our own multi-author documents by randomly selecting 10 to 50 sentences from source documents within the same genre. Specifically, we generate documents containing writings from two to four authors, with each author appearing exactly once within each document. 
We train 2-layer MLP classifiers using DeBERTa and LUAR-Mistral embeddings respectively. For the extended context variant of Mistral, we set the window size to \(k = 4\) and employ a 3-layer MLP classifier. 
We used an 80/20 train/test split within the same genre.

\begin{table}[]
\begin{tabular}{l|l|ll} \hline

\textbf{Genre}            & \textbf{DeBERTa} & \textbf{Mistral} & \textbf{+k-sent} \\ \hline
\textbf{B.G. Reviews}        & 0.710            & 0.731               & \textbf{0.734}            \\
\textbf{Instruction}      & 0.609            & 0.686               & \textbf{0.694}            \\
\textbf{journalism}       & \textbf{0.809}            & 0.780               & 0.798            \\
\textbf{LIT. forum} & 0.776            & \textbf{0.790}               & 0.779            \\
\textbf{STEM forum}       & \textbf{0.913}            & 0.744               & 0.755        \\\hline   
\end{tabular}
\caption{F1 scores on the five HRS genres. Notations: B.G.- Boardgame, LIT. - Literature.}
\label{tab:HRS_result}
\end{table}

Table \ref{tab:HRS_result} shows that DeBERTa and LUAR-Mistral perform better on different document genres. DeBERTa significantly outperforms on the STEM forum, where the discussion is conversational and features domain-specific terminology and slang, similar to PAN-25. Mistral significantly outperforms on instructional documents, where the writing format is highly standardized, with clear, step-by-step instructions, and pre-defined technical terms and verbs are used. This makes it harder for style change detection. This is also the hardest HRS set that has the lowest F1 for both models, however Mistral's degradation is less than DEBERTa's. Otherwise, DeBERTa slightly outperforms on journalism, which focuses on reporting specific events, while Mistral performs better on boardgames reviews and literature forum, which tend to be more subjective and expressive. 


\subsection{System Combination}

We concatenate DeBERTa and LUAR-Mistral embeddings to train the classifier. 
The reason for such a combination is to leverage the characteristics of bi-encoder transformer-based features and those of forward LLM-based features to produce a more generalized multi-author transition detection capability.
It is a simple system combination method and does not require large-scale modifications to the underlying feature extractors. 
We trained the following configurations: (1) 2-MLP layer classifier trained on DeBERTa features concatenated with sentence-level LUAR-Mistral features, (2) 3-MLP layer classifier trained on DeBERTa features concatenated with multi-sentence-level LUAR-Mistral features. Results are presented in the lower section of Table \ref{tab:pan2526_results}. 
The combined system achieves competitive performance against standalone benchmarks, even outperforming them on the PAN-25 and PAN-26 medium and hard tasks.

\section{Conclusion}

We explore cross-domain adaptation of an LLM using the LUAR framework for authorship attribution. Experiments show that such LLM embeddings can capture stylistic information for multi-author detection, and using multiple-sentence features yields better detection than single-sentence features. We also experimented with multiple embedding models across various types of user-generated text, and proposed a more all-rounded system by combining DeBERTa and LLM embeddings for effective style change detection.



\section{Limitations}

A key limitation of pre-trained LLMs is the lack of control over their internal data, making it difficult to verify if our experimental datasets were included in their training. Additionally, the response time of the combined system remains slow due to the integration of multiple features and use of LLM, even though LoRA improves the robustness of the Mistral model. 

Our initial experiments reveal that the decoder-based model and the encoder-based model excel in different genres, which suggests the possibility that these models are capturing different linguistic dimensions. More in-depth study is needed to validate this hypothesis and determine the reasons behind these variations. 
A more detailed study on style-change datasets, especially machine generated text, is also needed to better understand the domain-dependent requirements for effectively evaluating stylistic features. Current multi-author detection datasets fail to reflect collaborative scenarios where authors intend to produce cohesive writing with smooth transitions. Further work is required to bridge the gap with real-world use cases.

\section{Acknowledgment}
The authors would like to thank the BBN ALERT team and the NLP reading group for the useful discussions. 
This research is based upon work supported in part by the Office of the Director of National Intelligence (ODNI), Intelligence Advanced Research Projects Activity (IARPA), via 2022-22072200003. The views and conclusions contained herein are those of
the authors and should not be interpreted as necessarily representing the official policies, either expressed or implied, of ODNI, IARPA, or the U.S. Government. The U.S. Government is authorized to reproduce and distribute reprints for governmental purposes notwithstanding any copyright annotation therein.

\bibliography{anthology}

\appendix

\label{sec:appendix}

\section{System design and decision}
\label{sec:app_decision}
\subsection{List of datasets to train LUAR-Mistral}

We trained our LUAR-Mistral with opensource datasets listed below:

\begin{itemize}
\item Social Media Platform:
\begin{itemize}
    \item Reddit \cite{khan2021deep, baumgartner2020pushshift}
    \item Twitter \cite{rangel2017overview, stamatatos2018overview, rangel2019overview}
\end{itemize}
\item Reviews and Comments:
\begin{itemize}
    \item IMDB \cite{seroussi2014authorship}
    \item Yelp \cite{yelp_dataset_2021}, Amazon \cite{ni2019justifying}
    \item New York Times \cite{nyt_comments_kaggle}
\end{itemize}
\item Formal Communication:
\begin{itemize}
    \item Email \cite{klimt2004enron}
\end{itemize}
\item Narratives and Stories:
\begin{itemize}
    \item Blog \cite{schler2006effects}
    \item Fan-Fiction \cite{kestemont2020overview}
    \end{itemize}
\end{itemize}

\section{Analysis of PAN-25 and PAN-26 Data}

\label{sec:appendix_dataset}


Table \ref{tab:dataset_info} presents a overview of the PAN-25/26 datasets, on the number of documents, sentences and average word per sentence. This gives an idea of how much data the system is exposed in training, and how different tasks deviate from each other. Notice the PAN-26 medium task is much larger than the other tasks, but surprisingly with lower F1 overall. 

We also list a sample for each of the task and provide a brief analysis on the document characteristics in Tables \ref{tab:pan25_easy}, \ref{tab:pan25_medium}, \ref{tab:pan25_hard}, \ref{tab:pan26_easy}, \ref{tab:pan26_medium}, \ref{tab:pan26_hard}. The transition sentences are highlighted in blue.
\begin{table}[h]
\begin{tabular}{lcccc} \hline
\multicolumn{5}{c}{\textbf{PAN-25}}                             \\ \hline
\multicolumn{1}{c}{Task} & Set   & \# Doc & \#Sent & \#Word/Sent \\\hline
Easy                     & Train & 8.4k   & 48k    & 18.31       \\
                         & Valid & 1.8k   & 10k    & 18.42       \\
                         & Test  & 1.8k   & 10k    & 18.55       \\\hline
Medium                   & Train & 8.4k   & 59k    & 18.91       \\
                         & Valid & 1.8k   & 13k    & 19.00       \\
                         & Test  & 1.8k   & 13k    & 18.90       \\ \hline
Hard                     & Train & 8.4k   & 51k    & 20.07       \\
                         & Valid & 1.8k   & 11k    & 20.49       \\
                         & Test  & 1.8k   & 11k    & 19.87       \\ \hline
\multicolumn{5}{c}{\textbf{PAN-26}}                             \\ \hline
\multicolumn{1}{c}{Task} & Set   & \# Doc & \#Sent & \#Word/Sent \\\hline
Easy                     & Train & 21k    & 549k   & 29.90       \\
                         & Valid & 4.5k   & 117k   & 29.84       \\\hline
Medium                   & Train & 21k    & 1299k  & 27.03       \\
                         & Valid & 4.5k   & 279k   & 27.01       \\\hline
Hard                     & Train & 21k    & 590k   & 28.01       \\
                         & Valid & 4.5k   & 127k   & 27.87      \\\hline
\end{tabular}
\caption{Detail of the train, valid and test sets for the 3 tasks in PAN25 and PAN26, presented in number of documents, sentences and average number of word in each sentence.}
\label{tab:dataset_info}
\end{table}


\begin{table*}[ht]
\begin{tabular}{p{15.5cm}} \hline
\multicolumn{1}{c}{\textbf{PAN-25 Easy (4 authors)}}    \\\hline In general, be courteous to others.\\ Debate/discuss/argue the merits of ideas, don't attack people.\\ Personal insults, shill or troll accusations, hate speech, any suggestion or support of harm, violence, or death, and other rule violations can result in a permanent ban.\\ \textcolor{blue}{Correcto.}\\ They refer to them more easily as animals or a disease.\\ Only suitable to be wiped out and systematically destroyed.\\ Like stray dogs with rabies.\\ \textcolor{blue}{r/politics is currently accepting new moderator applications.}\\ If you want to help make this community a better place, consider !\\\textcolor{blue}{Exactly.}\\ The response should have been to impound every bus, arrest, charge, and hold every driver on human trafficking and kidnapping charges and issue a warrant for the arrest of Abbott and any staff of his or the bus lines identified as involved.\\ Yup.\\ Wilhoit's law: “Conservatism consists of exactly one proposition, to wit: There must be in-groups whom the law protects but does not bind, alongside out-groups whom the law binds but does not protect.”. \\ \hline
\end{tabular}
\caption{Problem 10 of PAN-25 easy training set. Beside the stylistic changes, transition can be identified easily with words that indicate the response to the previous users -- "Correct", "Exactly" and topic header -- "r/politics".}
\label{tab:pan25_easy}
\end{table*}
\begin{table*}[ht]
\begin{tabular}{p{15.5cm}} \hline

\multicolumn{1}{c}{\textbf{PAN-25 Medium (2 authors)}}   \\ \hline
Other players will get called up and the world cup is the biggest stage to present yourself.\\ Players get huge contracts after world cups because they played well.\\ It simply won't happen.\\ Major FAs like Brazil, Germany, France, England and so on could have joined forces and declined to play, that would have been an option.\\ But those are most of the time on the same level of corruption.\\ I don't blame the players, it would have been a Kaepernick situation just in a different sport.\\ Their name would have been dragged through the mud and FIFA would have done anything in their power to discredit them.\\ \textcolor{orange}{Football} is dirty behind the scenes and I don't think it's as easy to drop out of the worldcup as a player as you think without facing major repercussions.\\ \textcolor{blue}{And by "terrible situation", he specifically means 6,500 deaths so far and counting.}\\ For a ** \textcolor{orange}{soccer} match.\\ FIFA has blood on its hands.\\ All those workers were real people with real lives and families.\\ They were poor and brown, but they were regular ass human beings at the end of the day.\\ How the fuck could they let this happen?\\ Would they have ignored even 2 or 3 deaths in America or Canada?\\ Forget 6,500.\\ \hline
\end{tabular}
\caption{Problem 10 of PAN-25 medium training set. The transition sentence is not trivial to detect, but change of style is observed in neighboring sentences with the use of "soccer" and "football" (highlighted in orange manually) that describe the FIFA event.}
\label{tab:pan25_medium}
\end{table*}
\begin{table*}[ht]
\begin{tabular}{p{15.5cm}} \hline

\multicolumn{1}{c}{\textbf{PAN-25 Hard (4 authors)}}     \\ \hline
Soooo… according to the article, the Chinese army started scheduled training drills in the area this morning, and all of a sudden, fishermen saw a ufo?\\ The US just confirmed the others were balloons too.\\ Schumer said it on an interview on ABC this morning.\\ They haven’t blamed them on China though.\\ Not officially.\\ But he spun the announcements of the balloons into talking about China’s balloon spy program.\\ there are only so many times you can use the word balloon in a pentagon briefing before people start to laugh it off.\\ \textcolor{blue}{Ah the old "look we have spy balloons invading us too, so those other spy balloons couldn't have been ours."}\\ trick.\\ Of course they are forgetting that they already demanded their downed spy balloons returned to them, the other day.\\ \textcolor{blue}{Yeah they weren't the same as the PRC balloon, however they were still balloons.}\\ I don't see why the general would call them balloons if they weren't.\\ 
Already have read your article.\\ It states these objects are not similar to the original PRC balloon, however they are still balloons according to Schumer and the Canadian General.   \\ \hline                                                     
\end{tabular}
\caption{Problem 10 of PAN-25 hard training set. The topical difference is not obvious, however transition words such as "Ah", "Yeah" that indicate the response to the previous users is still presented in the transition sentence.}
\label{tab:pan25_hard}
\end{table*}

\begin{table*}[ht]
\begin{tabular}{p{15.5cm}}
\hline
\multicolumn{1}{c}{ \textbf{PAN-26 Easy (3 authors, 5 transitions)}} \\\hline
“Y/N Look out” You quickly turned your head only to see one of the Devaronians aim their blaster towards you.\\ Instead of answering, you glared daggers at him, which caused him to snicker as he made his way towards you.\\ \textcolor{blue}{A quick glance at your fuel reserves told you that would have ended in disaster, so, with a fake smile plastered on your face and a bowed head for anyone in uniform, you quickly explained that you were on a tour of First Order allied planets as a gift from your parents for your eighteenth birthday, and that you had been cruising around before realizing you would need to land as soon as possible to refuel.}\\ \textcolor{blue}{“Y/N, I… I haven’t been completely honest with you, and I need to tell you this before...” “W-what do you mean, Poe?}\\ “Poe—“ “Please, Y/N, let me finish or I will never get this out,” You stayed silent and Poe continued.\\ “You’re going to be a fun one to break.” As he finally released you, your gaze went straight down to Poe who continued to lay motionless on the ground.\\ And in that moment, as you looked up into Poe’s dark eyes, you knew that everything was going to be okay.\\ Y/N, are you okay?” You realized Poe was behind you and you tried to look back towards him, only to be able to see the side of his face through your peripherals.\\ “You really are somethin’, you know that Dameron?” Poe smiled at you in a way that you somehow knew was only reserved for you as he smoothed your Y/H/C out of your face, leaning forward to place a feather light kiss on your forehead.\\ “Just someone who is crazy about you, sweetheart.” You sighed happily and leaned into Poe’s touch—wrapping your arms to lessen the little distance that was between the two of you.\\ “Y/N… hang on sweetheart, I’m coming for you.” The dust cleared and you could see that it was Poe getting closer to you, his hand reaching out.\\ Didn’t I tell you I would come up with a plan to get us out of here?” You let out a breathy laugh and looked up to see Poe smiling slightly.\\ This is getting little inten—“ “Already ahead of you.” Poe waved the com-link in your face and you smirked.\\ \textcolor{blue}{The resulting explosions had you thinking that these were some sort of weapons manufacturers, and suddenly the knot in your stomach made sense; whatever this planet was used for obviously dealt in death.}\\ \textcolor{blue}{Tanned skin, dark curly hair, and warm brown eyes smile down at you while he holds out his hand for you to shake.}\\ \textcolor{blue}{You tried to rush into your ship, hoping to get on the comms and call out to the X-Wings that you were a friendly, but your ship was hit with a blast that sent you flying; unlike the Troopers that had been helping you, you got tossed to an open field – they collided with a nearby wall with a sickening crunch.}\\ \hline
\end{tabular}
\caption{PAN-26 Easy train set problem 10. Due to the nature of consistency in stroy telling, the change of topic is different than PAN-25 without transition words.}
\label{tab:pan26_easy}
\end{table*}

\begin{table*}[ht]
\begin{tabular}{p{15.5cm}}
\hline
\multicolumn{1}{c}{ \textbf{PAN-26 Medium (20/118 lines)}} \\\hline
Steve goes to Pepper for help because even though he's only just met her and he still doesn't know her very well, he can tell that she's a good person with a kind heart and that she won't say no to any reasonable request.\\ He also knows that she is the sort of person who can get things done and that she will do the things that need to be done with a minimum of fuss and bother.\\ They meet for coffee in Bryant Park on a sunny day in March and Pepper listens quietly as Steve outlines what he wants and why.\\ Pepper doesn't blink an eye, just whips her StarkPad out of her purse and starts tapping away as they brainstorm ideas.\\ Bucky is the most alive person Steve has ever known and the idea of putting his name on a grave is unimaginably wrong.\\ He immediately vetoes her suggestion of a memorial stone in Arlinon because he just cannot bear the thought of a tomb of any type.\\ In the end Steve decides that he doesn't want to plant a sapling and chooses one of the heritage trees; a big white oak that curves majestically toward the sun and that has a canopy as big as a small swimming pool.\\ When they hit upon the commemorative tree program in Prospect Park they both know that they've found the answer.\\ But a tree – a living thing that could grow and flourish for more than a life time - a tree is just perfect.\\ Steve does not deliberately seek out the tree, but he does sometimes glimpse it out of the corner of his eye if his daily run takes him that way.\\ But after he presses them for more information they estimate that it might have been planted sometime in the early twentieth century and that settles it in Steve's mind.\\ The park authorities aren't entirely sure how old it is because the records are inconclusive and unclear.\\ \textcolor{blue}{There are no signs anyone else has been down this way recently, and when they get to the first village, they are told that no one else has come down from the mountains in weeks.}\\ \textcolor{blue}{There are more bad days than good - tears intermingled with stony silence or fits of rage so vast they make a tsunami look tame.}\\ And every day he wakes up and hopes that this one will be better than the last, that he'll open his eyes to find that everything has shifted for the better.\\ Despite his not wanting to make a fuss and his desire to keep things as simple as possible, he's actually insanely pleased to see them troop through the door and settle at the table.\\ In late October they hold a simple ceremony to unveil the plaque, just Steve, Pepper and Tony in attendance.\\ Afterwards they go to a local steak house for prime rib and beer and somewhere around 3pm the other Avengers start to trickle in.\\ They end up taking over a whole corner of the restaurant, ordering more food and more beer and being generally themselves, which is actually kind of scary when Steve thinks about it closely enough.\\ But the owners can't really complain once Tony has slapped his credit card on the bar, ordering them to charge everything to him, and they stand back and let them carry on regardless.
\\... \\ \hline
\end{tabular}
\caption{Sub-document of PAN-26 Medium train set problem 10. Due to the nature of consistency in stroy telling, the change of topic is less obvious than PAN-25. But with careful reading we can notice the location changes from park (city) to village and mountain (remote) area at the adjacent transition sentences.}
\label{tab:pan26_medium}
\end{table*}

\begin{table*}[ht]
\begin{tabular}{p{15.5cm}}
\hline
\multicolumn{1}{c}{ \textbf{PAN-26 Hard (2 authors, 1 transition)}} \\\hline
Your tears flowed freely after that and you leaned forward to press your lips to his forehead, clinging lightly to him but avoiding the injuries from the battle. \\
You sniffed, not even realizing until then that you had begun to cry, as he tried to comfort you, tears of grief and guilt clinging to his lower lashes. \\
“You nearly died Thorin,” you whispered, turning your palms to cup his bearded jaw, “You nearly died and I almost didn’t get to tell you…”. \\
\textcolor{blue}{“Well now Thorin, what’s going on because I can tell that it’s not anything to do with an illness.” Oin was very matter of fact when it came to these things.} \\
“How did you know?” Thorin wasn’t startled by his friend and general, the dwarf did live in the same wing as the king, in fact most of the company did. \\
Thorin looked over at his general, his eyebrow raised in question, the gruff dwarf turned red and nodded. \\
She looked at Dwalin who nodded and took over for her, she nodded back and moved to follow Thorin as he walked out of the training area. \\
Dwalin turned to the woman behind him “Good on ya lass, mind staying for a moment?” she nodded, sheathing her weapons as he turned back to the rest of the new recruits. \\
She would roll with this but damn, (Y/N) glanced quickly over at Dwalin as she raised her hand in salute and stayed that way. \\
“Aye lass, what is it?” Dwalin was fairly friendly at least when he wasn't grumbling about something. \\
“Oh I’ve known where those two trouble makers have been and where they will be for the next few days as well” Dwalin chuckled. \\
“Aye, you and the rest of us” came Kili's response as he chuckled out “would have loved to see you get knocked on your ass at least once, Dwalin”. \\
“Dwalin” came Fili’s voice “Never thought I would see you get your ass handed to you by a woman no less, just wait till I tell Thorin” Fili’s laughter boomed across the training area. \\
“Be here tomorrow at dawn” Dwalin was finally able to get out, with that he clapped Fili on the shoulder and guided him out of the training area leaving (Y/N) to herself.
\\ \hline
\end{tabular}
\caption{Sub-document of PAN-26 Hard train set problem 11. Transition is harder to observe with the fluency of the story.}
\label{tab:pan26_hard}
\end{table*}

\end{document}